\documentclass[conference]{IEEEtran}
\IEEEoverridecommandlockouts

\usepackage{balance}

\usepackage{cite}

\ifCLASSINFOpdf
  \usepackage[pdftex]{graphicx}
  \DeclareGraphicsExtensions{.pdf,.jpeg,.png,.jpg}
\else
\fi

\usepackage{url}

\begin{document}

\newcommand\blfootnote[1]{%
  \begingroup
  \renewcommand\thefootnote{}\footnote{#1}%
  \addtocounter{footnote}{-1}%
  \endgroup
}

\title{CEASEFIRE: An AI-Powered System for Combating Illicit Firearms Trafficking \thanks{Research was possible due to the funding from the European Union’s
Horizon Europe research and innovation programme under grant agreement
No 101073876 (Ceasefire).}}

\author{
    \IEEEauthorblockN{
        Jorgen Cani\textsuperscript{1},
        Ioannis Mademlis\textsuperscript{1}, 
        Marina Mancuso\textsuperscript{2},
        Caterina Paternoster\textsuperscript{2},
        Emmanouil Adamakis\textsuperscript{3}, \\
        George Margetis\textsuperscript{3},
        Sylvie Chambon\textsuperscript{4},
        Alain Crouzil\textsuperscript{4},
        Loubna Lechelek\textsuperscript{4},
        Georgia Dede\textsuperscript{5},
        Spyridon Evangelatos\textsuperscript{5},\\
        George Lalas\textsuperscript{5},
        Franck Mignet\textsuperscript{6},
        Pantelis Linardatos\textsuperscript{7},
        Konstantinos Kentrotis\textsuperscript{7},
        Henryk Gierszal\textsuperscript{8},
        Piotr Tyczka\textsuperscript{8},\\
        Sophia Karagiorgou\textsuperscript{9},
        George Pantelis\textsuperscript{9},
        Georgios Stavropoulos\textsuperscript{10},
        Konstantinos Votis\textsuperscript{10} and
        Georgios Th. Papadopoulos\textsuperscript{1}\\
    }
    
    \IEEEauthorblockA{
        \small
        \textsuperscript{1}Department of Informatics and Telematics, Harokopio University of Athens,
            E-mail: \{cani, imademlis, g.th.papadopoulos\}@hua.gr \\
        \textsuperscript{2}Università Cattolica del Sacro Cuore – Transcrime,
            E-mail: \{marina.mancuso, caterina.paternoster\}@unicatt.it \\
        \textsuperscript{3}Foundation for Research and Technology Hellas, Institute of Computer Science,
            E-mail: \{madamakis, gmarget\}@ics.forth.gr \\
        \textsuperscript{4}IRIT, Université de Toulouse, CNRS,
            E-mail: \{sylvie.chambon, alain.crouzil, loubna.lechelek\}@irit.fr \\
        \textsuperscript{5}Netcompany-Intrasoft,
            E-mail: \{georgia.dede@netcompany.com, spyros.evangelatos, george.lalas\}@netcompany.com \\
        \textsuperscript{6}Thales,
            E-mail: franck.mignet@nl.thalesgroup.com \\
        \textsuperscript{7}EXUS Software Single Member Limited,
            E-mail: \{p.linardatos, k.kentrotis\}@exus.ai \\
        \textsuperscript{8}ITTI,
            E-mail: \{henryk.gierszal, piotr.tyczka\}@itti.com.pl \\
        \textsuperscript{9}Ubitech Ltd,
            E-mail: \{skargiorgou, gpantelis\}@ubitech.eu \\
        \textsuperscript{10}Centre for Research and Technology Hellas,
            E-mail: \{stavrop, kvotis\}@iti.gr \\
    }
}

\maketitle


\begin{abstract}
Modern technologies have enabled illicit firearms trafficking to partially merge with cybercrime, while also allowing its off-line aspects to become increasingly complex. The online trade of firearms, their components, 3D blueprints and illicit substances carried out by criminals on both the surface Web and dark Web is increasingly difficult to address as a consequence of the exponential growth in the amount of information disseminated on the Internet. On the other hand, law enforcement agencies are confronted with significant challenges that require the development of sophisticated technological solutions capable of processing large volumes of data, identifying relevant information in a timely manner and creating networks of connections between potential criminal groups. This article presents a real-world practical system, namely the CEASEFIRE one, powered by advanced artificial intelligence technologies that can assist law enforcement personnel in addressing the above described challenges.
\end{abstract}


\section{Introduction}


Trafficking of illicit firearms is one of the main sources of revenue for organized crime groups, while providing them the main tools for their antisocial trade: guns. Today’s digital economy and breakneck technological progress have made it even more difficult for state authorities to properly control the situation. Law Enforcement Agencies (LEAs) face difficult challenges that were not even conceivable just a few years ago.

For instance, criminals thrive in Dark Web marketplaces, which are hidden illegal e-shops that require specific URLs to access and constantly change locations to evade law enforcement. These platforms, along with cryptocurrency transactions, enable anonymous exchanges of illegal goods like firearms and drugs, complicating LEA efforts. Meanwhile, the Surface Web and social media are used by firearms traffickers for information exchange and coordination, where the vast volume of data hampers detection. Advances in 3D printing have fostered an on-line community sharing weapon blueprints, attracting criminal interest. Additionally, criminals exploit postal services to smuggle firearms, taking advantage of legal discrepancies between countries and the massive volume of daily parcels that conceal illicit goods. Lastly, identifying and matching seized firearms to international databases is challenging due to diverse weapon appearances, potential serial number tampering, and insufficient training among law enforcement practitioners.

In short, modern technologies have led illicit firearms trafficking to partially merge with advanced cybercrime, while simultaneously permitting its off-line aspects to become more sophisticated as well. Artificial Intelligence (AI) holds the promise of helping LEA practitioners to surpass these challenges, by automating much of the grunt work in which they regularly engage. The adoption of state-of-the-art AI technology and tools by state authorities is very likely to lead to enormous LEA productivity increases, compared to the recent past. \cite{mademlis2023invisible}

\textbf{CEASEFIRE} (``Advanced versatile artificial intelligence technologies and interconnected cross-sectoral fully-operational national focal points for combating illicit firearms trafficking")\blfootnote{Project Website: \url{https://ceasefire-project.eu/}} tries to achieve exactly that. Funded by the European Commission's Horizon Europe programme, it is a 3-year innovation project that began in October 2022 and unites the expertise of 7 companies, 5 research organizations, and 9 law enforcement agencies (LEAs) across Europe. The CEASEFIRE project is developing a cutting-edge technical system that integrates innovative AI tools and algorithms to enhance the capabilities of LEAs in combating firearms trafficking and related criminal activities more efficiently. Its objective is to comprehensively and effectively address all of the aforementioned challenges, by leveraging state-of-the-art technology. This article reviews the specific pressing issues that the CEASEFIRE system is designed to face via its use-cases and subsequently provides a broad overview of its technical design and components. These serve as an example of how modern technological advancements can be harnessed to bolster law enforcement efforts and curb the proliferation of illegal firearms.

\section{Related Work}
This ambitious vision of comprehensive AI-powered systems for fighting illicit goods trafficking has begun to take shape in various studies, but typically using limited data sources and modalities. For instance, individual systems use a custom ontology and Knowledge Graph (KG) to aid criminal investigations and generate admissible digital evidence, but they focus solely on social media content \cite{Elezaj2020}, on-line newspapers \cite{Srinivasa2019}, or semi-automatic Web crawling \cite{Kejriwal2019}. In contrast, the system presented in \cite{Carrillo2020} uses diverse data sources like official crime reports and socioeconomic information, manually entered into a KG, to uncover criminal patterns, specifically in Mexico City. Other approaches aim to use even more varied data sources but restrict their scope to specific criminal cases. Among the most comprehensive prototypes are those that analyze the Surface Web, Dark Web, and cryptocurrency transactions, identifying links between various illegal activities \cite{Mazzonello2021}, and others that gather data from multiple sources, including social networks, financial data, and geospatial data \cite{Muller2022}.

These trends showcase the need for a comprehensive system that exploits state-of-the-art technologies and all available data sources, in order to exhaustively analyze the complex multi-faceted trafficking phenomenon.

\section{The CEASEFIRE Use-Cases}
\label{sec::UseCases}
The CEASEFIRE project is building exactly such a system with advanced AI and ICT components, focusing on 5 real-world use-cases relevant to firearms trafficking.

\textbf{Use-Case \#1}: \textit{Real-time systematic firearms incident and intelligence information collection and exchange}.\\
The effective collection, correlation, and sharing of timely, accurate intelligence information from multiple sources is crucial for analyzing and investigating firearms trafficking. However, Europe's current system is fragmented with inconsistent, imprecise data and varied data formats. Developing a user-friendly tracking tool for firearm incidents would enhance information exchange and aid law enforcement in criminal investigations. As a result, the first goal of CEASEFIRE is to develop a near-real-time European-scale firearms incident tracking tool that will also support the efficient intelligence information exchange among the various (types of) LEAs.

\textbf{Use-Case \#2}: \textit{On-the-spot firearm seizure registration and cross-border data search}.\\
During police investigations, seized firearms must be accurately identified and matched against databases to gather critical information. This task is challenging due to variations in appearance, the need to know specific details like caliber size and serial number locations, and alterations by criminals. Law enforcement officers typically lack specialized training for this precise identification. Developing automatic visual-based analysis tools for use on-site and on smartphones would greatly assist officers in this work. Therefore, the second goal of the CEASEFIRE system is to develop a mobile application to automatically and on-the-spot identify the main characteristics (e.g., brand, model, caliber, location of the serial number, etc.) of a seized firearm at the crime scene, in order to subsequently facilitate automatic search/update of information in relevant (inter-)national databases. The respective mobile application is to be directly used on-site by police officers at crime scenes.

\textbf{Use-Case \#3}: \textit{Firearms purchase on Dark Web marketplaces}.\\
Criminal actors are progressively exploiting the multiple advantages offered by the Internet, by taking the opportunity to expand their portfolio of products/services and, also, by fragmenting their business over a range of on-line monikers and marketplaces. As a result, Dark Web markets represent one of the most commonly used platforms for the trade of illicit firearms and constitutes a priority threat for law enforcement activity. Consequently, the third goal of the CEASEFIRE system is to develop tools for automatic analysis of on-line activities, in order to detect and reveal the real identities of individuals engaged in illegal firearms trafficking on darknet marketplaces.

\textbf{Use-Case \#4}: \textit{Mail order and courier service firearms trafficking detection using scanning technologies}.\\
Criminals increasingly use postal and courier services to traffic firearms, components, and ammunition within the EU, exploiting legislative gaps between countries. Firearms or parts restricted in one country can be easily ordered on-line from another, using fast parcel delivery services. As a result, the fourth CEASEFIRE goal is to develop a tool for automatically detecting illicit firearms, ammunition or firearms components in X-ray scan images of parcels, which have been mailed with legit post and courier services. The respective application is to be directly used on-site by officers at customs offices equipped with X-ray scanners.

\textbf{Use-Case \#5}: \textit{3D-printed firearm blueprints distribution}.\\
An international on-line community of 3D gun-printing enthusiasts has emerged, anonymously sharing blueprints, advice, training materials, and building networks. This decentralized, leaderless setup makes the community difficult to track and stop. They communicate across multiple digital platforms, including Twitter and IRC. Therefore, the fifth CEASEFIRE goal is to develop tools for automatically detecting and analyzing the on-line distribution of blueprints of 3D-printed firearms.

\section{The CEASEFIRE System}
\label{sec::System}
The ultimate goal of the CEASEFIRE system is to introduce a greatly innovative, high-tech and versatile approach to tackle trans-border firearms trafficking incidents, by developing and integrating advanced AI solutions for boosting the LEA practitioners’ everyday work. To this end, a toolkit of innovative, robust, efficient AI-enabled technologies is being developed, which will significantly boost the practitioners’ effectiveness and productivity. 

This system is informed by efforts to reliably handle current gaps, limitations, malfunctions and inefficiencies in establishing fully-functional National Firearms Focal Points (NFFPs), i.e., designated bodies in each EU member state that coordinate efforts to combat illegal firearms trafficking, provide guidance on EU firearms regulations, facilitate cooperation and information exchange among LEAs, and serving as the primary liaison with EU institutions on firearms-related issues. Due to inherent interdisciplinary aspects and considerations, the CEASEFIRE system's technical design and components fundamentally rely on a concurrent joint effort to:
\begin{itemize}
\item Analyze relevant criminal modi operandi, develop crime commitment risk profiles, and identify connections with other forms of organized crime.
\item Define the NFFPs organizational chart, advance transborder and cross-jurisdictional cooperation between agencies from multiple
countries and/or of different type.
\item Coordinate with the activities of the EMPACT-Firearms operational action plans and their corresponding specific operational actions. The EMPACT-Firearms security initiative\footnote{\url{https://www.europol.europa.eu/crime-areas-and-statistics/empact}} is an EU-led program aimed at tackling illegal firearms trafficking through enhanced cooperation and intelligence sharing among EU member states. 
\item Harmonize the relevant legal framework regarding the investigation procedures, the adherence to the applicable legislation, and the improvement of the current regulation to keep up with the recent criminal developments.
\item Define forensics protocols for lawful court-proof collection, processing and exchange of crime evidence.
\end{itemize}

\begin{figure*}
\label{fig::FunctionalArchitecture}
\centerline{\includegraphics[width=\linewidth]{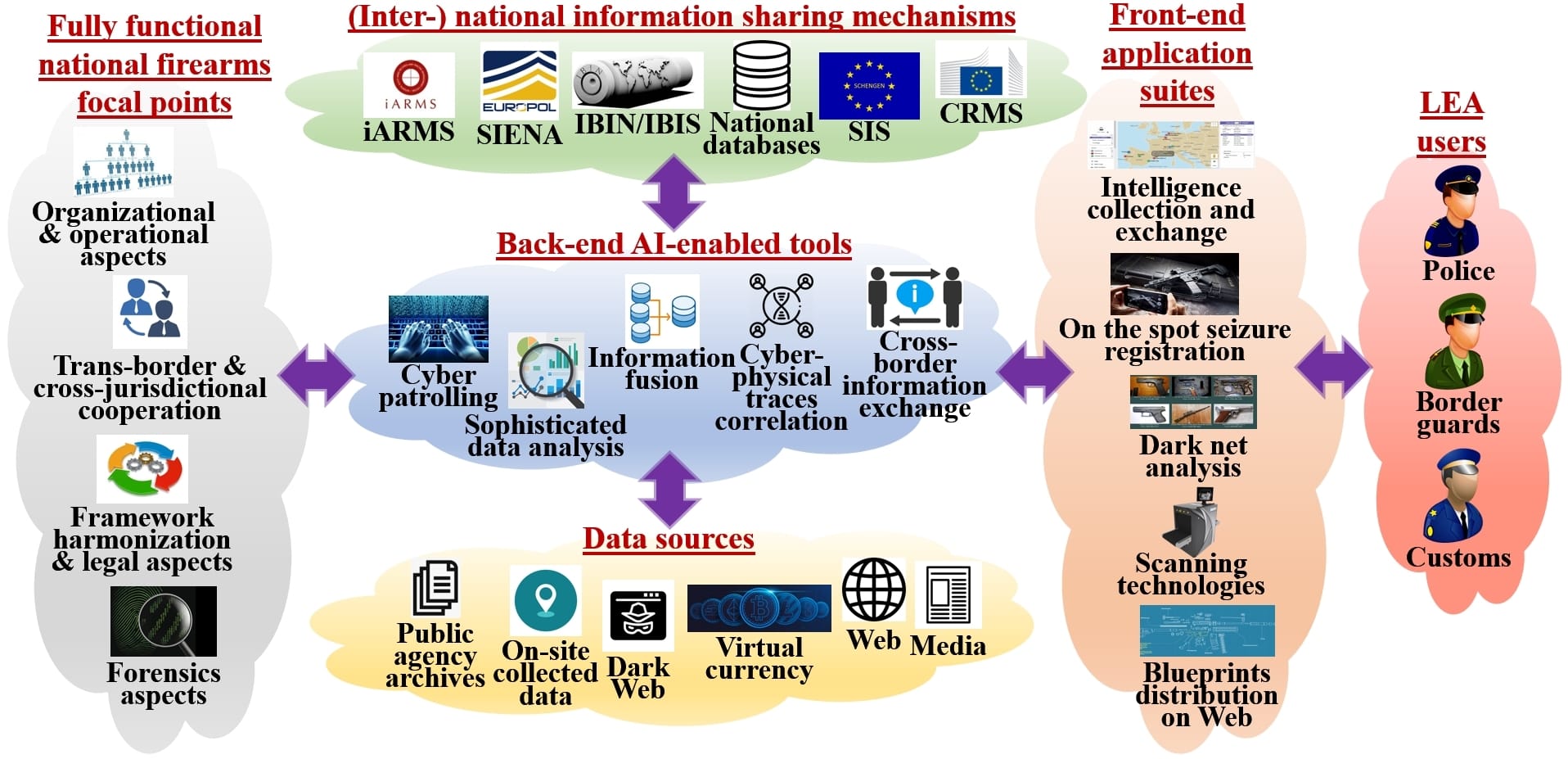}}
\caption{The CEASEFIRE system's functional architecture.}\vspace*{-5pt}
\end{figure*}

Built on top of these underlying efforts, the CEASEFIRE system supports the following groups of AI technologies:
\begin{itemize}
\item \textbf{Cyber patrolling}, which covers aspects ranging from open source monitoring and dark net marketplace examination to incident data gathering and cryptocurrency analysis. These target to successfully access and monitor on-line sources.
\item \textbf{Sophisticated data analysis}, which involves tools for processing all types of collected multimedia data (e.g., textual data, visual content, etc.). These aim at efficiently analyzing the typically enormous amounts of information gathered from on-line sources or data obtained on-site and discovering interesting evidence information.
\item \textbf{Information fusion}, which incorporates a set of large-scale data-driven methods for realizing the fusion of all different information sources and the robust detection of patterns in the collected data.
\item \textbf{Cyberphysical traces correlation}, which involves a combination of machine-learning-based and semantic technologies for performing the actual linking between the collected cyber/digital traces and the real identities of the perpetrators/suspects.
\item \textbf{Cross-border information exchange}, which concerns the development of the necessary underlying infrastructure to enable the efficient access of the system to multiple multiagency international databases as well as the effective exchange of information among the LEA practitioners.
\end{itemize}

In order to implement the vision of CEASEFIRE, a wide set of heterogeneous data sources needs to be analyzed, covering the cases of both on-line and off-line investigation activities. These data sources are provided as input to the back-end AI-enabled modules described above, while the overall goal is to support near-real-time analysis and generation of actionable intelligence. In particular, the CEASEFIRE system supports the following list of input data sources:
\begin{itemize}
\item \textbf{Public agency archives}. This concerns archival data that are maintained by the various LEA agencies and which can serve as enabling factors for developing the required analytics technologies.
\item \textbf{On-site collected data}. This corresponds to data that are captured on-the-spot during the practitioners’ investigation attempts (e.g., images of the seized firearms at the crime scene, scanned images of parcels, etc.).
\item \textbf{Dark Web}. This refers to data collected from the unindexed parts of the Web, such as dark nets and darknet marketplaces.
\item \textbf{Cryptocurrency transactions}. This refers to the transaction graphs of cryptocurrency systems (e.g., Bitcoin, Ethereum, etc.).
\item \textbf{Surface Web}. This concerns open information sources that are typically used for suspicious communication activities (e.g., exchange of training materials, guidance on constructing firearms, information on the existing legal gaps, etc.), including social media, blogs, forums, repositories, etc.
\item \textbf{Media}. This corresponds to publicly available information that is being communicated by verified sources (e.g., news articles).
\end{itemize}

A critical dimension in any attempt to construct a firearms trafficking handling system, due to the inherently transnational and multifaceted nature of the problem at hand, is to develop efficient, user-friendly and interoperable communication interfaces with current (inter-)national information sharing mechanisms. Thus, CEASEFIRE supports the interconnection of various information sharing mechanisms related to firearms trafficking, including:
\begin{itemize}
\item Interpol’s ``Illicit Arms Records and tracing Management System (iARMS)", whose  primary goal is to facilitate firearms tracing, by maintaining a world-wide record of seized firearms.
\item Europol’s ``Secure Information Exchange Network Application (SIENA)", which comprises a platform for facilitating the swift exchange of operational and strategic crime-related information among LEA agencies.
\item Interpol’s ``Ballistic Information Network (IBIN)", whose target is to facilitate the sharing of ballistics data and the identification of links among gun-related crimes at a global scale.
\item ``Schengen Information System (SIS)", which comprises the largest information sharing system for security and border management in Europe and which enables competent national LEA authorities to exchange intelligence and to enter/consult alerts on incidents of interest.
\item ``Common Customs Risk Management System (CRMS)", which aims at serving as a fast and easy-to-use mechanism to exchange risk-related information directly between operational officials and risk analysis centers in the EU.
\item National databases, where the exchange of information with transnational peers is critical.
\end{itemize}

The CEASEFIRE system aims to provide user-friendly front-end applications for LEA practitioners, in direct accordance to the aforementioned use-cases, to enhance their operational capabilities using AI-enabled back-end tools. Designed for interoperability, it supports police, border guards, and customs authorities across borders, ensuring smooth information exchange with judicial authorities and forensics experts.

\begin{figure*}
\label{fig::TechnicalApproach}
\centerline{\includegraphics[width=\linewidth]{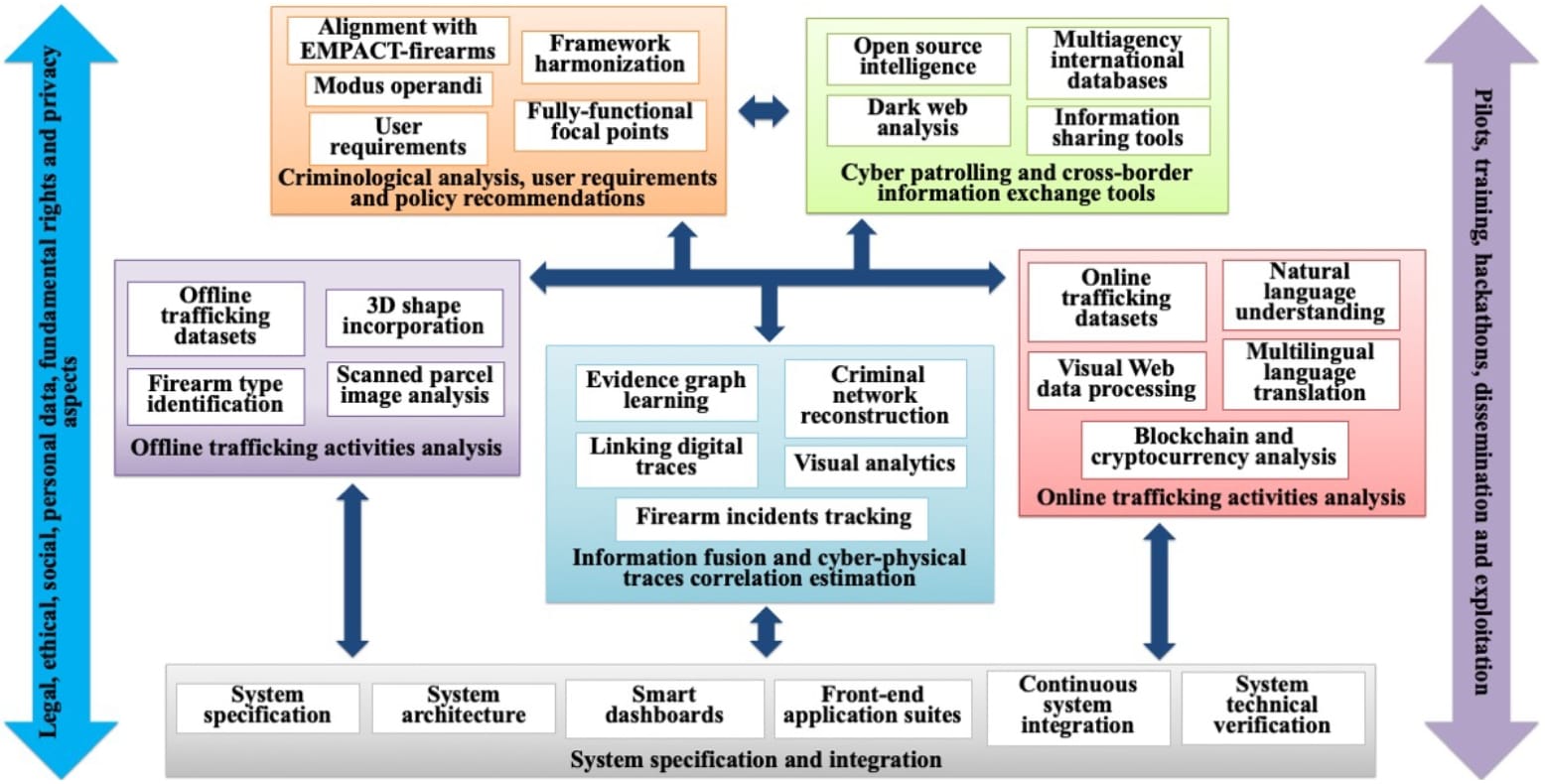}}
\caption{The CEASEFIRE system's technical approach.}\vspace*{-5pt}
\end{figure*}

The system is organized according to a structured approach of multiple layers, each one composed of multiple innovative components. This technical approach contains the following layers:

\textbf{Cyber-patrolling and cross-border information exchange tools}: This layer is responsible for realizing data collection from Web sources and exchange of information among LEA agencies. It involves tools for (semi-)automated open-source intelligence collection and real-time systematic firearms incident data gathering, analysis of Dark Web and darknet marketplaces, interconnection of multi-agency international databases to allow simultaneous inputs/queries, as well as efficient information sharing tools among LEA, judicial and forensics authorities. This layer directly addresses the needs of Use-Cases \#3 and \#5.

\textbf{Off-line trafficking activities analysis}: This layer handles all data processing pipelines for off-line trafficking activities. It houses the creation of relevant datasets, in order to facilitate the development of relevant AI modules, as well as AI-powered tools for firearm detection in X-ray scan images of mailed parcels and for RGB-based on-site firearm type identification and seizure registration. This layer directly addresses the needs of Use-Cases \#2 and \#4.

\textbf{On-line trafficking activities analysis}: This layer implements all data processing pipelines for on-line trafficking activities. It foresees the creation of relevant datasets for facilitating the development of relevant AI modules, as well as various AI-powered tools for automatic analysis of data crawled from the Surface/Dark Web, so as to automatically detect content related to firearms trafficking. These include auxiliary services for multilingual speech-to-text and language translation to convert textual content into English, as well as tools for image analysis, for natural language understanding in order to semantically parse in-depth discussions available in on-line forums and blogs, or for blockchain and cryptocurrency analysis, in order to detect patterns, correlations and other critical information cues in transactional graphs for various commonly met types of cryptocurrencies (e.g., Bitcoin, Ethereum, etc.). This layer directly addresses the needs of Use-Cases \#3 and \#5.

\textbf{Information fusion and cyberphysical traces correlation estimation}: This layer realizes information fusion and estimates correlations among the cyber and the physical worlds. It contains data-driven AI-powered tools for automatic evidence graph learning and firearms-related key event detection, by analyzing and fusing the various types of heterogeneous/multimodal data sources, for linking digital traces with physical entities via sophisticated data mining and correlation estimation approaches, for reconstruction and analysis of organized criminal networks, by applying advanced analytics and causal modeling methods, as well as for evidence-based visual analytics of cyberphysical traces. This layer also houses a tool for near-real-time pan-European incident tracking, which supports the search and analysis of firearms-related incidents and directly addresses the needs of Use-Case \#1.

\section{AI for combating firearms trafficking}
Based on the presented functional architecture and technical approach, the CEASEFIRE system architecture has been developed according to the logical view depicted in Figure 3. The architecture is presented in a layered pattern, with four vertical and four horizontal layers, briefly described below.

\begin{figure*}
\label{fig::LogicalView}
\centerline{\includegraphics[width=\linewidth]{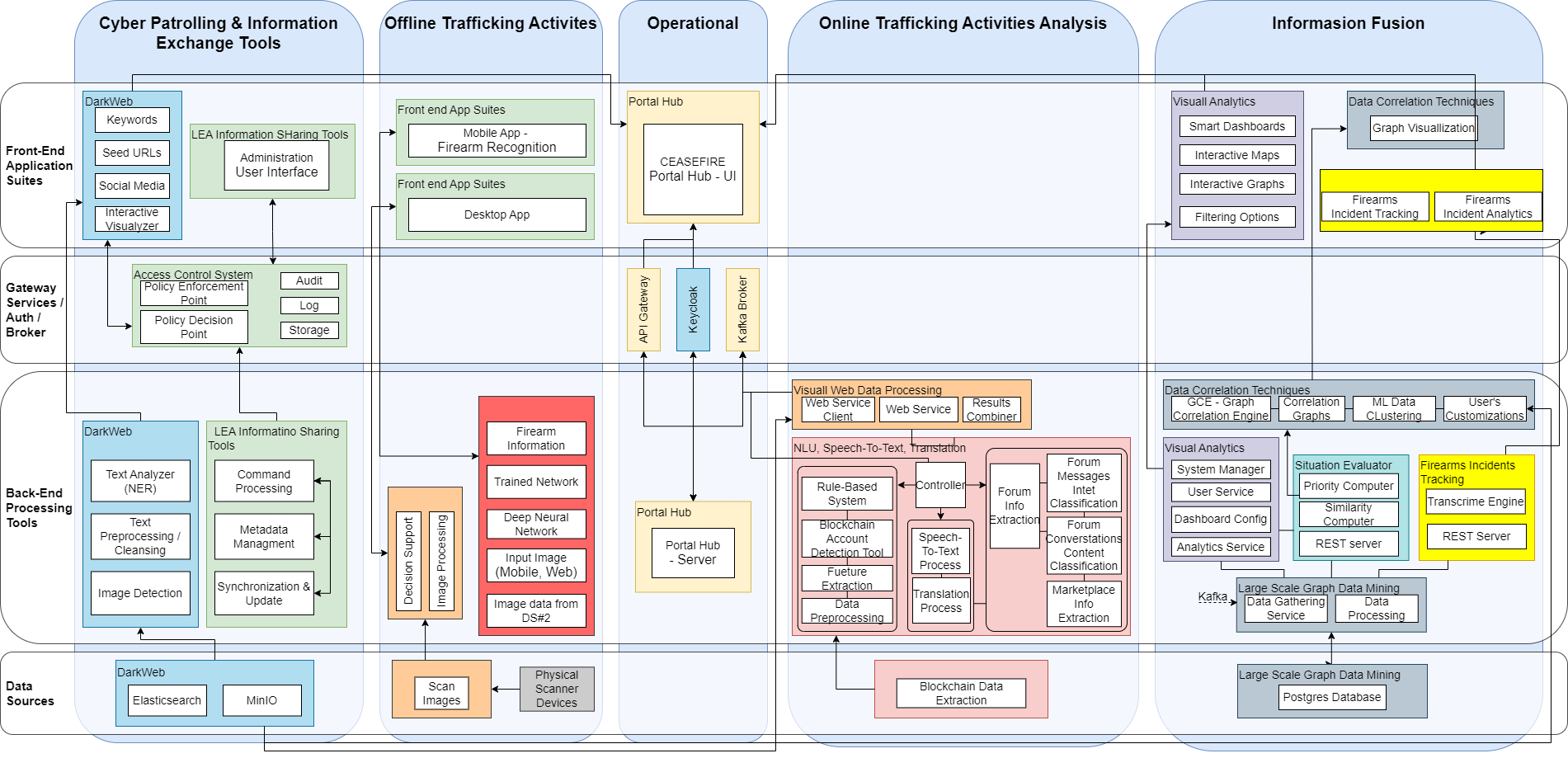}}
\caption{The CEASEFIRE system architecture's logical view.}\vspace*{-5pt}
\end{figure*}

In summary, each vertical layer groups the system building components according to application functionality.
\begin{itemize}
\item \textbf{Cyber-patrolling and cross-border information tools}: Layer responsible for realizing data collection from Web sources and exchange of information across LEAs.
\item \textbf{Off-line Trafficking Analysis}: Layer responsible for handling all data pipelines for off-line trafficking activities.
\item \textbf{On-line Trafficking Analysis}: Layer implementing all data processing pipelines for on-line trafficking activities.
\item \textbf{Information Fusion and cyberphysical traces correlation estimation}: Layer realizing information fusion and estimating correlations among the cyber and the physical worlds.
\item \textbf{Operational}: This is an added layer, in which the components responsible for the smooth operation of the CEASEFIRE platform reside.
\end{itemize}

The horizontal layers group system modules according to their role in data processing.
\begin{itemize}
\item \textbf{Front-end application suites}: This layer contains all the inner-components modules of the services that will provide a UI environment, so that the user can interact with the provided services and generate reports. These UIs are be hosted in the central Portal Hub, either as integrated applications or with an external link.
\item \textbf{Gateway services}: This layer contains the mechanisms used by every service to communicate and exchange information. It’s the place where authentication, routing, load balancing, and API interactions are managed.
\item \textbf{Back-end processing tools}: This layer contains all the inner-component modules with the actual technical implementation of the systems logic, such as AI tools, text/image processing and language translation modules.
\item \textbf{Data sources}: This layer encompasses the various repositories and databases that store information valuable for every component.
\end{itemize}

In general, common technical solutions are utilized for integrating the various system components and providing inter-module communication, such as RESTful Web Services and the respective clients, Apache Kafka for data ingestion, etc. The software assets of the entire system are gathered into the CEASEFIRE \textbf{Portal Hub}, to be easily accessible by the LEA end-users. Its core functionalities revolve around giving access to the CEASEFIRE services and making any produced reports available to the LEA officers. Thus, the Portal Hub offers a comprehensive User Interface where the user, after successfully authenticating, gains an overview of all the CEASEFIRE services that are available to them. After selecting the desired service, the user gets redirected to it without the need of repeating the authentication process.

Additionally, through the Portal Hub, the user has access to a ``Reports Overview" page where he/she can manage all the reports created across the CEASEFIRE services. The user can search among all the reports, while utilizing a variety of filters (e.g., creation date) or he/she can choose to install, delete or share a report with another user. The back-end of the tools, also, exposes endpoints, giving the ability to the partners (service provider) to manage how their asset is portrayed inside the Portal.

The actual innovations and state-of-the-art functionalities of the system are located in the back-end services. A brief presentation of the most important ones is given below.
\subsection{Cyber-patrolling and cross-border information exchange tools}
\textbf{Dark Web analysis}. This asset provides a semantically aware crawling service for navigation in Dark Web forums and darknet marketplaces, which exposes a graphical user interface (GUI) to LEA officers. The crawling system utilizes multiple technologies:
\begin{itemize}
\item TOR\footnote{\url{https://www.torproject.org/}}, a piece of software that facilitates communication with the Dark Web and anonymous navigation in the Surface Web.
\item MongoDB\footnote{\url{https://www.mongodb.com/}}, as the in-memory database of the system that holds the static metadata of the crawling process.
\item MinIO\footnote{\url{https://min.io/}}, as an object store where all the images from the Surface/Dark Web are being stored for further analysis.
\item ElasticSearch\footnote{\url{https://www.elastic.co/}}, i.e., a distributed search and analytics engine where all the gathered data and relevant metadata are kept.
\item Natural Language Understanding (NLU) AI models for textual analytics, identifying and extracting entities related to the CEASEFIRE context.
\item Named Entity Recognition (NER) AI models, facilitated by a continuously updated list of slang/special terms for text analysis, in order to to identify specific keywords in a text and classify them as entities related to firearms discussions or illegal bitcoin transactions.
\end{itemize}

The GUI can be used to specify the desired target to crawl, keywords, see the status of the process, browse among the collected evidence, text, images, or video, and view the analysis results of the performed crawl. As a result, on top of the collected data, LEA officers can perform descriptive queries and derive high level analytics.

\textbf{LEA Information Sharing Tools}. This asset provides real-time aggregation and federation of data coming from multi-agency international databases, thus providing the LEA users the ability to interact with multiple and diverse databases with a single request. Mechanisms are included to manage potential operations between databases, such as access control, security policies, and user role definitions. From a technology perspective, this asset relies on SQL databases and Trino\footnote{\url{https://trino.io/}}, i.e., a distributed SQL query engine for big data analytics.

LEA users are able to share information about firearms between LEAs’ databases as well as between LEA database and the CEASEFIRE one. The scope of information exchange, source and destiny of datasets, as well as control properties of the exchange process is managed by the CEASEFIRE administrator, following LEA requirements.
\subsection{Off-line Trafficking Analysis}
\textbf{Scanned parcel image analysis}. This asset is a back-end service that allows automatic detection of firearms components/ammunition in X-ray scans of parcels. It relies on a appropriately pretrained Deep Neural Network (DNN), composed of a Convolutional Neural Network (CNN) backbone and a Vision Transformer head for object detection \cite{Cani2024}. The performance of the pipeline is further improved by pretraining the DNN on an unlabeled set of data, using self-supervised training methods \cite{konstantakos2024self}, particularily useful on low-data settings such as the X-ray illicit firearms domain. The back-end service communicates directly with the respective front-end Web application with a suitable GUI for customs officers. The application pushes to the service requests containing X-ray images and, for each image, receives back the type of object detected and an on-image bounding box (in pixel coordinates). Overall, 10 object classes are supported (e.g., `Revolver', `Shotgun', etc.). After each analysis session, a brief report in the form of a JSON file can be optionally sent to the CEASEFIRE database. This asset powers the CEASEFIRE tool developed for the system's Use-Case \#4.

\textbf{Firearm type identification}. This asset is a back-end service that allows recognition of the model of a firearm in an RGB photograph, taken by a mobile phone at a crime scene. It relies on an appropriately pretrained Vision Transformer DNN for fine-grained visual classification, which incorporates a background subtraction module \cite{Chou2023}. The utilisation of diffusion models \cite{alimisis2024advances} enables the generation of synthetic images of firearms for use by the DNN in its training phase, thereby enhancing the visual classification rate. The back-end service communicates directly with the respective front-end mobile application that supplies the photographs to be analyzed and receives back the 5 most likely types of weapons per image. After each analysis session, a brief report can be optionally sent to the CEASEFIRE database. This asset powers the CEASEFIRE tools developed for the system's Use-Case \#2.

\subsection{On-line Trafficking Analysis}
A back-end ``Controller" mechanism regulates the AI-powered automated analysis of data gathered from Surface/Dark Web forums or marketplaces, by exploiting the following modules:

\textbf{Visual Web data processing}. This is an asset for recognizing entities related to firearms trafficking in RGB images crawled from the Web. It integrates a client that sends requests with a service that receives them and provides a composite image recognition functionality. The service operates via an ensemble of two, appropriately trained DNNs that combine CNN and Vision Transformer components. The first DNN exploits self-supervised pretraining before domain-specific downstream finetuning, to accurately recognize 23 different entities (e.g., bomb, rifle, drug powder, etc.) in unoccluded images \cite{Konstantakos2023}. The second one supports only a subset of this class set, but is robust even to heavy occlusions or noise \cite{Kortylewski2020}. The predictions of both DNNs are combined via an appropriate rule system, taking into account their known accuracy, and the final output is returned to the client.

\textbf{Natural Language Understanding} (NLU). This is a composite asset that accurately detects semantic entities and events related to firearms trafficking in textual sources, including on-line forums and blogs. It uses information extraction and classification methods to identify suspicious elements, such as listing specifications, transactions, and detailed information about firearm-related activities. The asset relies on an HTML-based rule system, which analyses the existing HTML tags of a marketplace page, as well as Transformer DNNs for question answering \cite{Liu2019}, to extract firearm and ammo specifications from an unstructured listing description, and Named Entity Recognition (NER), to extract listing information even in severely unstructured cases.

Regarding forums, it relies on two Transformer DNNs finetuned for text classification, which identify conversations as relevant/irrelevant (relevance model) and as potentially suspicious/non-suspicious (suspiciousness model). Irrelevant dialogues are related to non-real firearms, for example firearms related to on-line games, historical events, etc., while relevant conversations concern hunting, gun laws, off-line and on-line firearm commerce, etc. Suspiciousness refers to conversations about firearm offers, requests, exchanges, undetected use, etc. An additional Transformer DNN finetuned for intent recognition further categorizes the messages in on-line dialogues \cite{Makrynioti2023} \cite{Bunk2020}. Finally, the potentially suspicious dialogues that have been attached a relevant intent (e.g., offer, exchange, request, etc.), are fed into a separate module for forum information extraction that distills precise information from the conversation threads, using NER and text summarization.

\textbf{Multilingual speech-to-text and translation}. This asset analyzes and translates audio and text data across multiple languages, using publicly available, pretrained Transformer DNNs. Audio signals are first transformed into textual format using speech-to-text tools. Subsequently, in the case of non-English text, it is translated into English from a variety of supported languages, utilizing multilingual translation technologies after the source language has been automatically detected. The resulting translated text is forwarded to the NLU asset for analysis.

\textbf{Blockchain and cryptocurrency analysis}. This asset analyzes Bitcoin and Ethereum cryptocurrency transactions to detect suspicious activities within blockchain networks. Commencing with the extraction and preprocessing of transaction data from the blockchains, including sender and receiver addresses, transaction amounts and timestamps, duplicate removal, normalization and transformation of addresses into functional data (via Functional Data Analysis, or FDA) are subsequently employed to facilitate efficient analysis. Advanced feature extraction methods are then employed, such as clustering and detection-controlled estimation methods based on transaction frequency, USD value, addresses' lifespan, or graph neighborhood features for accurate address classification, in order to identify patterns indicative of illicit activities relevant to firearms trafficking.

In the next step, an Long Short-Term Memory (LSTM) neural network and other machine learning classification models (Random Forest, eXtreme Gradient Boosting) are applied to forecast primary activities of blockchain addresses, by analyzing balance trends over time. Finally, a rule system that exploits a carefully organized database of suspicious users automatically flags transactions based on predefined criteria, reinforcing the identification of suspicious behaviors and establishing correlations between addresses in Ethereum, created concurrently with the inactivity of corresponding Bitcoin addresses \cite{Blitsi2023}. The algorithm checks timestamps and firearm prices to establish connections between illicit addresses from both Bitcoin and Ethereum networks. The application of FDA for address activity prediction enables addresses to be categorized into distinct groups based on primary activities, operating independently of network-related information.

\subsection{Information Fusion and cyberphysical traces correlation estimation}
Despite being prominent in contexts such as analysis of illicit Dark Web transactions or 3D-printed firearm blueprints distribution, this vertical layer of the system's logical view is inherently designed to be Use-Case-agnostic. It provides a versatile platform that can be adapted to various investigative scenarios encountered by LEAs, by fusing heterogeneous information to provide a unified view of the criminal landscape, enabling analysts to evaluate specific crime reports, as well as assisting them in understanding whether a firearms trafficking-related crime shares connections with similar events, or different types of criminal acts. Additionally, this layer houses the tools implementing the system's Use-Case \#1.

This layer's back-end components are the following ones:

\textbf{Large-scale graph data mining}. This asset aims at synthesizing and structuring vast datasets derived from the other vertical layers of the system's logical view into a comprehensive graph database. It involves real-time aggregation of varied data streams, leveraging Apache Kafka for data ingestion and meticulously preparing the data through cleansing, feature engineering, and encoding. The resulting PostgreSQL database is optimized for advanced graph data mining, setting the stage for powerful analytics and automated extraction of insights.

\textbf{Graph Correlation Engine} (GCE). This asset extracts and visualizes complex correlations from a multitude of data sources, primarily focused on entities related to firearms. GCE implements two primary methods: a custom scoring algorithm for initial correlation identification, and a clustering algorithm (DBSCAN). The first one parses through data to identify potential correlations based on predefined criteria, enabling users to discern patterns and linkages between disparate pieces of information. DBSCAN clustering complements this by evaluating the spatial density of the data to autonomously configure the number of clusters, effectively grouping entities without pre-setting the cluster count. These entities, which could be individuals (agents) or events with connections to firearms, are characterized by identifiable attributes such as name or username, date, country of origin, and specific firearm details. To streamline the analysis, GCE simplifies the data by grouping dates into quarters and years, categorizing countries by continents, and classifying firearms into a set of the top 10 types. This level of abstraction not only aids in the reduction of complexity, but also enhances the interpretability of the correlations and clusters, thereby providing LEAs with actionable insights and a macro-level understanding of firearm-related dynamics.

\textbf{Situation evaluator}. This asset computes a priority for each event (e.g., an offering for sale of firearms on a dark net) and a similarity between events, based on a model that incorporates criminology domain knowledge of traffickers' modus operandi. The underlying modeling methodology is a Bayesian Network that is able to capture uncertainties given scarce data \cite{Taroni2014}. The asset also offers a service that responds to user queries for events: in the front-end application, the user can specify certain criminal traces and ask for ``similar" events, in the sense that they could have caused the observation.

\textbf{Visual analytics}. This asset provides the back-end software for a Use-Case-agnostic front-end application that allows visualization and interactive exploration of illegal firearms trafficking trails. To achieve this, information and data collected from the rest of the CEASEFIRE system are firstly properly structured using a suitable semantic ontology and database schema, after data cleaning, normalization, and integration. The application presents to LEA officers all available data sources and identified correlations for analysis, providing access to the records of each data source, as well as search and filtering capabilities. Officers can add pertinent data source records to a strategic intelligence case related to their analytical objective. Subsequently, they can review the cases and, based on their judgment, either continue the analysis or generate a report to take further action.

\textbf{Firearm incidents tracking}. This asset is the back-end component of the near-real-time European-scale firearms-related incidents tracking tool developed for the system's Use-Case \#1. It contains an engine that first gathers on-line news articles related to specific firearm incidents, i.e., seizures, homicides, shootings and armed robberies, before processing this sources in order to identify eligible articles according to defined criteria. Subsequently, a standard set of relevant information (such as incident type, date, location, firearm involved, number of victims, number of perpetrators, etc.) is automatically extracted from these articles by applying NER and information retrieval methods, in order to develop risk indicators, identify red flags and conduct data analysis.

\section{Conclusion}
The latest high-tech trends used by organized crime for firearms trafficking can be effectively countered by employing specialized versions of cutting-edge AI methods for large-scale data processing. This is crucial given the growing digital complexity of trafficking networks and the significant harm their activities cause, whether social, economic, or environmental. Operational integration of such AI tools into LEA procedures provides a practical solution to address the current criminal threats. The presented CEASEFIRE system serves as a comprehensive example of addressing these concerns.

Building on the significant advancements made by the CEASEFIRE project in tackling trans-border firearms trafficking, future research directions can further enhance LEA effectiveness. One promising area is the development of advanced predictive analytics and machine learning models, which can forecast potential trafficking hotspots and analyze behavioral patterns of known offenders to enable preemptive actions. Enhancing real-time data integration by incorporating multisource data fusion and blockchain technology for data integrity will create a more comprehensive and secure situational awareness framework, which can be assisted by AI methods for data-driven vulnerability assessment and automatic risk indicators generation. Cybersecurity measures can be bolstered by sophisticated Dark Web monitoring tools and AI-driven cyber forensic techniques, which can trace the origins and networks of illegal transactions. To address the growing threat of 3D-printed firearms, developing advanced algorithms to detect encrypted blueprints and infiltrating decentralized networks will be crucial. The platform can be extended to support in the fight against other types of organized crime as well, besides firearms trafficking. Establishing a global intelligence-sharing framework supported by AI tools, along with advocating for legal harmonization across countries, will close loopholes exploited by traffickers. Finally, focusing on human factors, such as immersive VR and AR training simulations for LEA officers and refining user-centered design interfaces, will ensure the practical effectiveness of these technologies. To this end, a Digital Twin solution can be integrated for more powerful visualization and user interaction, while the entire system can be implemented following a security-as-a-service conceptualization.

%
\IEEEpeerreviewmaketitle



\balance
\bibliography{refs}
\bibliographystyle{ieeetr}

\end{document}